# LitMC-BERT: transformer-based multi-label classification of biomedical literature with an application on COVID-19 literature curation

Qingyu Chen, Jingcheng Du, Alexis Allot, and Zhiyong Lu

**Abstract**— The rapid growth of biomedical literature poses a significant challenge for curation and interpretation. This has become more evident during the COVID-19 pandemic. LitCovid, a literature database of COVID-19 related papers in PubMed, has accumulated over 180,000 articles with millions of accesses. Approximately 10,000 new articles are added to LitCovid every month. A main curation task in LitCovid is topic annotation where an article is assigned with up to eight topics, e.g., Treatment and Diagnosis. The annotated topics have been widely used both in LitCovid (e.g., accounting for ~18% of total uses) and downstream studies such as network generation. However, it has been a primary curation bottleneck due to the nature of the task and the rapid literature growth. This study proposes LITMC-BERT, a transformer-based multi-label classification method in biomedical literature. It uses a shared transformer backbone for all the labels while also captures label-specific features and the correlations between label pairs. We compare LITMC-BERT with three baseline models on two datasets. Its micro-F1 and instance-based F1 are 5% and 4% higher than the current best results, respectively, and only requires ~18% of the inference time than the Binary BERT baseline. The related datasets and models are available via https://github.com/ncbi/ml-transformer.

———————————— ◆ ————————————

## 1 INTRODUCTION

The rapid growth of biomedical literature significantly challenges manual curation and interpretation [1, 2]. These challenges have become more evident under the context of the COVID-19 pandemic. Specifically, the median acceptance time of COVID-19 papers is about 2-time and 20-time faster than the acceptance time for papers about Ebola and cardiovascular disease [3]. The number of articles in the literature related to COVID-19 is growing by about 10,000 articles per month [4]. LitCovid [5, 6], a literature database of COVID-19 related papers in PubMed, has accumulated a total of more than 100,000 articles, with millions of accesses each month by users worldwide. LitCovid is updated daily, and this rapid growth significantly increases the burden of manual curation. In particular, annotating each article with up to eight possible topics, e.g., Treatment and Diagnosis, has been a bottleneck in the LitCovid curation pipeline. Fig. 1. shows the characteristics of topic annotations in LitCovid; we will explain the related curation process below. The annotated topics have been used both in LitCovid directly and downstream studies widely. For instance, topic-related searching and browsing account for over 18% of LitCovid user behaviors [6], and the topics have also been used downstream studies such as citation analysis and knowledge network generation. application [7-9]. Therefore, it is important to develop automatic methods to overcome this issue.

Innovative text mining tools have been developed to facilitate biomedical literature curation for over two decades [2, 10-12]. Topic annotation in LitCovid is a standard multi-label classification task, which aims to assign one or more labels to each article [13]. To facilitate manual topic annotation, we previously employed the deep learning model Bidirectional Encoder Representations from Transformers (BERT) [14]. We used one BERT model per topic, known as Binary BERT, and previously demonstrated this method achieved the best performance of the available models for LitCovid topic annotations [6]; other studies have reported consistent results [15]. Indeed, existing studies in biomedical text mining have demonstrated Binary BERT achieves the best performance in multi-label classification tasks [16, 17]. However, this method has two primary limitations. First, by training each topic individually, the model ignores the correlation between topics, especially for topics that often co-occur, biasing the predictions and reducing generalization capability. Second, using eight models significantly increases the inference time, causing the daily curation in LitCovid to require significant computational resources.

To this end, this paper proposes a LITMC-BERT, a transformer-based multi-label classification method in

————————————————

- *Qingyu Chen is with the National Center for Biotechnology Information (NCBI), National Library of Medicine (NLM), National Institutes of Health (NIH), 9000 Rockville Pike, Bethesda, Maryland 20892. E-mail: qingyu.chen@nih.gov.*
- *Jingcheng Du is School of Biomedical Informatics, UT Health, 7000 Fannin St #1200, Houston, TX 77030. E-mail: Jingcheng.Du@uth.tmc.edu.*
- *Alexis Allot is with the National Center for Biotechnology Information (NCBI), National Library of Medicine (NLM), National Institutes of Health (NIH), 9000 Rockville Pike, Bethesda, Maryland 20892. E-mail: alexis.allot@nih.gov.*
- *Zhiyong Lu is with the National Center for Biotechnology Information (NCBI), National Library of Medicine (NLM), National Institutes of Health (NIH), 9000 Rockville Pike, Bethesda, Maryland 20892. E-mail: zhiyong.lu@nih.gov.*

*Corresponding author: Zhiyong Lu*

biomedical literature. It uses a shared BERT backbone for all the labels while also captures label-specific features and the correlations between label pairs. It also leverages multi-task training and label-specific fine-tuning to further improve the label prediction performance. We compared LITMC-BERT to three baseline methods using two sets of evaluation metrics (label-based and example-based) commonly used for multi-label classification on two datasets: the LitCovid dataset consists of over 30,000 articles and the HoC (the Hallmarks of Cancers) dataset consists of over 1,500 articles (the only benchmark dataset for multi-label classification methods in biomedical text mining).. It achieved the highest performance on both datasets: its instance-based F1 and accuracy are 3 and 8% higher than the BERT baselines, respectively. Importantly, it also achieves the SOTA performance in the HoC dataset: its micro-F1 and instance-based F1 are 5% and 4% higher than the current best results reported in the literature, respectively. In addition, it requires only ~15% of the inference time needed for Binary BERT, significantly improving the inference efficiency. LITMC-BERT has been employed in the LitCovid production system, making the curation more sustainable. We also make the related datasets, codes, and models available via https://github.com/ncbi/ml-transformer.

## 2 RELATED BACKGROUND

### 2.1 Multi-label classification

Multi-label classification is a standard machine learning task where each instance is assigned with one or more labels. Multi-label classification methods can be categorized into two broad groups [18]: problem transformation, transforming the multi-label classification problem into relatively simpler problems such as single-label classification and algorithm adaptation, adapting the methods (such as changing the loss function) for multi-label data.

The methods under the problem transformation group are traditional approaches to address multi-label classification. Most popular methods include (1) binary relevance, where each label requires to train a corresponding binary classification model [19], (2) label powerset, where a binary classification model is trained for every combination of the labels [20], and (3) classifier chains, where the output of a binary classification model is used as the input to train a further binary classification model [21]. Such methods have achieved promising performance in a range of multi-label classification tasks [13, 22]. Indeed, existing studies have shown binary relevance BERT achieved the best performance for topic annotation in LitCovid [6, 15]. However, it is computationally expensive and transforming multi-label classification tasks into binary classification may ignore the correlations among labels. Recently, an increasing number of deep learning methods under the algorithm adaptation group have been proposed which predict all the labels directly as the output [23-25].

### 2.1.1 Multi-label text classification in the domain of biomedical text mining

Text classification methods have been widely applied in biomedical text mining for biomedical document triage

[26], retrieval [27], and curation [28]. Compared to the general domain, text classification especially for multi-label classification in biomedical text mining has three primary challenges: (1) domain-specific language ambiguity, e.g., a gene may have over 10 different synonyms mentioned in the literature; conversely, a gene and a chemical could share the same name [29]; (2) limited benchmark datasets for method development and validation; e.g., the HoC dataset [30] is the only multi-label text dataset among commonly-used benchmark datasets for biomedical text mining [16, 17] and it only has about 1,500 PubMed documents; and (3) deployment difficulty, i.e., an important contribution of biomedical text mining is to make open-source tools and servers such that biomedical researchers can readily apply. Therefore, the designed methods should be scalable to massive biomedical literature and are also efficient in standard research tool production settings where computational resources are limited such as the graphics processing unit (GPU) is not commonly available for method deployment.

Such challenges impact the method development in biomedical text mining. Most of the existing methods focus on transfer learning which employs word embeddings (such as WordVec) or transformer-based models (such as BERT) that are pre-trained in biomedical corpora to extract text representations [31-34]. Indeed, this also applies to the multi-label classification methods in biomedical text mining. Existing studies have mostly used Binary BERT [16, 17] for multi-label classification: for each label, it trains a corresponding BERT (or other types of transformers) classification model. The evaluation results show that the BERT-related approaches achieve the best performance for multi-label classification in biomedical text mining compared to other multi-label classification methods that have been used in the general domain [35, 36].

### 2.2 LitCovid curation pipeline

A primary focus of this study is to develop a multi-label classification method to facilitate COVID-19 literature curation such as topic annotation in LitCovid. Here we summarize the topic annotation in the LitCovid curation pipeline and its challenges.

The detailed LitCovid curation pipeline is summarized in [6]. For topic annotation, an article in LitCovid is considered for one or more of eight topics (General information, Mechanism, Transmission, Diagnosis, Treatment, Prevention, Case report, or Epidemic forecasting) when applicable. Fig. 1. shows the characteristics of topic annotations of the articles in LitCovid by the end of September 2021. Prevention, Treatment, and Diagnosis are the topics with the highest frequency. Over 20% of the articles have more than one topic annotated. Some topics co-occur frequently, such as Treatment and Mechanism, where papers describe underlying biological pathways and potential COVID-19 treatment (https://www.ncbi.nlm.nih.gov/pubmed/33638460). The annotated topics have been demonstrated to be effective for information retrieval and have been used in many downstream applications. Specifically, topic-related searching and browsing account for over 18% of LitCovid user behaviors among millions of accesses and it has been the second most accessed features in LitCovid [6]. The topics have also been used downstream studies such as evidence attribution, literature influence analysis, and knowledge network generation. application [7-9].

However, annotating these topics has been a primary bottleneck for manual curation. First, compared to the general domain, biomedical literature has domain-specific ambiguities and difficulties of understanding its semantics. For example, the Treatment topic can be described in different ways including patient outcomes (e.g., 'these factors may impact in guiding the success of vaccines and clinical outcomes in COVID-19 infections'), biological pathways (e.g., 'virus-specific host responses and vRNA-associated proteins that variously promote or restrict viral infection), and biological entities (e.g., 'unique ATP-binding pockets on NTD/CTD may offer promising targets'). Second, compared to other curation tasks in LitCovid (document triage and entity recognition), topic annotation is more difficult due to the nature of the task (assigning up to eight topics) and the ambiguity of natural languages (such as different ways to describe COVID-19 treatment procedures). Initially, the annotation was done manually by two curators with little machine assistance. To keep up with the rapid growth of COVID-19 literature, Binary BERT has been developed to support manual annotation. However, previous evaluations show that it has an F1-score of 10% lower than the tools assisting other curation tasks in LitCovid [6]. Also, Binary BERT requires a significant amount of inference time because each label needs a separate BERT model for prediction. This challenges the LitCovid curation pipeline, which may have thousands of articles to curate within a day.

# 3. DATA AND METHOD

## 3.1 Experiment datasets

We used LitCovid BioCreative and HoC datasets for method development and evaluation. TABLE 1 provides the characteristics.

For the LitCovid BioCreative dataset [37], it contains 24,960, 6,239, and 2,500 PubMed articles in the training, development, and testing sets, respectively. The topics were assigned using the above annotation approach consistently. All the articles contain both titles and abstracts available in PubMed and have been manually reviewed by two curators. The only difference is that the datasets do not contain the General Information topic since the priority of the topic annotation is given to the articles with abstracts available in PubMed [6]. In addition, the testing set contains the articles that have been added to LitCovid from 16th June to 22nd August after the construction of the training and development datasets. Using incoming articles to generate the testing set will facilitate the evaluation of the generalization capability of automatic methods. To our knowledge, this dataset is one of the largest multi-label classification datasets on biomedical English scientific text.. We have made this dataset publicly available to the community via https://ftp.ncbi.nlm.nih.gov/pub/lu/LitCovid/biocreative/. For the HoC dataset, it contains 1,580 PubMed abstracts with 10 currently known hallmarks of cancer that were annotated by two curators. The data set is available via https://www.cl.cam.ac.uk/~sb895/HoC.html. We used the same dataset split from previous studies [16, 24]; the training, development, and testing sets contain 1,108, 157, and 315 articles, respectively. As mentioned, it is the only dataset used for multi-label classification in biomedical literature from commonly-used benchmark datasets [16, 17, 24].

## 3.2 LITMC-BERT architecture

The architecture of LITMC-BERT is summarized in Fig. 2 and Fig. 3. Fig. 2 compares its architecture with other approaches (we will also use them as baselines), whereas Fig. 3 details its underlying modules. The detailed hyperparameters are also summarized in TABLE 2 and 3.4.1.

As mentioned, most biomedical text mining studies have used Binary BERT (Fig. 2 (A)) for multi-label classification [16, 17]. Indeed, we have applied Binary BERT to annotate topics in the LitCovid curation pipeline as well [6]. An alternative approach is to uses a shared BERT model with a Sigmoid function (or other similar activation functions) to outputs all the label probabilities directly (Fig. 2 (B)), which we denote it as Linear BERT (Fig. 2 (B)): it uses a Sigmoid function (or other similar activation functions) followed by a shared BERT model which outputs all the label probabilities directly. Linear BERT also forms the basis of LITMC-BERT (Fig. 2 (C)). In contrast, for LITMC-BERT, each label has its own module (Label Module) to capture label-specific representations; and the label representations are further used (Label Pair Module) to predict whether a pair of labels co-occurs. It also leverages multi-task training and label-based fine-tuning. We explain each in detail below.

### 3.2.1 Transformer Backbone

The Transformer Backbone applies a transformer-based model to get a general representation of an input text; in this case, the input text is the title and abstract (if available) of an article. In this study, the transformer-based model is BioBERT [38], which a BERT model pre-trained on PubMed and PMC articles. We evaluated a range of BERT variants and BioBERT (v1.0) gave the overall highest performance as the backbone model.

### 3.2.2 Label Module

Each label has a label module to capture its specific representations for the final label classification. Figure 3 (A) shows its detail. Essentially, the Label Module combines the final hidden vector for the CLS token of a BERT backbone (Figure 3 (A)(1); we call it CLS vector) and the label-specific vector (Figure 3 (A)(2)) to produce the final label feature vector (Figure 3 (A)(5)).

Using the CLS vector of a BERT backbone is recommended by the authors of BERT for classification tasks [14]. For LITMC-BERT, it is shared by all the labels (since a shared BERT model is used as the backbone). In addition, for each label, a Multi-head Self-Attention [39] and a global average pooling layer are applied to the last encoder layer of the BERT backbone (Figure 3 (A)(2)) to get a label-specific vector. This is designed to capture specific features for each label. We further normalize the CLS vector and label-specific vector with a multi-layer perceptron (MLP) consisting of a few dense layers (Figure 3 (A)(3)) (Figure 3 (A)(4)). This approach has been demonstrated to be effective for combining feature vectors from different sources [28]. The normalized vectors are summed up to produce the final label vector (Figure 3 (A)(5)).

### 3.2.3 Label Pair Module

The Label Pair Module further uses the label representations from the Label Module and captures correlations between label pairs. Figure 3 (B) shows its detail.

For a pair of labels 1 and 2, the Label Pair Module first uses their corresponding feature representations produced by the Multi-head Self-Attention in the Label Module (Figure (A)) as inputs. Then it performs co-attentions (Figure 3 (B)(2)) and global average pooling (Figure 3 (B)(3)) to get two vectors from the inputs. The co-attention mechanism is an adaption of the Multi-head Self-Attention whereas the query and key components of the Self-Attention are the label pairs in this case (e.g., the attention from label 1 to label 2 and the attention from label 2 to label 1 in Figure 3 (B)(2)). This has been demonstrated to be effective for modeling correlations between pairs [40, 41]. Then, the two vectors are fused using the same method above (Figure 3 (B)(4) and Figure 3 (B)(5)) to get the final label pair vector (Figure 3 (B)(6)). The label pair vector is used to predict whether the labels 1 and 2 co-occur as auxiliary tasks for the multi-training process introduced below. Auxiliary tasks are not directly related to primary tasks (label predictions in this case) but have shown effective for multi-task training to make the shared representation more generalizable [42]. In addition, while the relations between label pairs are important, it does not necessarily apply to every label pair. We define a hyperparameter called label pair threshold: the Label Pair Module is only applied to a label pair if above the threshold. For a pair of labels 1 and 2, the threshold is calculated by the number of instances that labels 1 and 2 co-occur dividing by the minimum number between the number of instances of labels 1 and 2 in the training set.

### 3.2.4 Multi-task training and label-based fine-tuning

The LITMC-BERT training process employs multi-task training where it trains and predicts the labels (main tasks) and co-occurrence (auxiliary tasks) simultaneously. The loss during the multi-task training is the total loss of main tasks and auxiliary tasks). Given that main tasks are the focus, we define a hyperparameter called auxiliary task weight (from 0 to 1) which takes a proportion of auxiliary task losses. The full hyperparameters and baselines are provided below. When the multi-task training converges, it further fine-tunes the Label Module for each label while freezing the weights of other modules . Such training approach has been shown effective in both text mining and computer vision applications [43, 44].

### 3.4 Baseline models

We compared LITMC-BERT to three baseline models: ML-Net (a shallow deep learning multi-label classification model which has achieved superior performance in biomedical literature) [24], Binary BERT (Fig. 2 (A)), and Linear BERT (Fig. 2 (B)).

ML-Net is an end-to-end deep learning framework which has achieved favorably state of the art (SOTA) performance in a few biomedical multi-label text classification tasks [24]. ML-Net first maps texts into high dimensional vectors through deep contextualized word representations (ELMo) [33], and then combines a label prediction network and label count prediction to infer an optimal set of labels for each document.

Binary BERT and Linear BERT are introduced in 3.2. For N labels, Binary BERT trains N BERT classification models (one label each) whereas Linear BERT provides all the N label predictions in one model. Note that previous studies mostly have used Binary BERT in biomedical literature [16, 17]. It also has been the state-of-the-art (SOTA) model for multi-label classification in biomedical literature [15-17] and was also used in the LitCovid production system previously [6].

### 3.4.1 Hyperparameters

For each model, we performed hyperparameter tuning on the datasets and selected the best sets of hyperparameter based on the validation set loss. TABLE 2 provides the hyperparameter values in the LitCovid BioCreative dataset; the configuration files of the hyperparameters are also provided in the github repository. Importantly, for BERT-related models (Binary BERT, Linear BERT, and LITMC-BERT), we controlled their shared hyperparameters (BERT backbone, maximum sequence length, learning rate, early stop steps, and batch size) to ensure a fair and direct comparison.

### 3.5 Evaluation metrics and reporting standard

There are a number of evaluation measures for multi-label classification tasks [13, 45, 46], which can be broadly divided into two groups: (1) label-based measures, which evaluate the classifier's performance on each label and (2) example-based measures (also called instance-based measures), which aim to evaluate the multi-label classifier's performance on each test instance. Both groups complement each other: in the case of topic annotation, label-based measures quantify the specific performance for each topic, whereas example-based measures quantify the effectiveness of models at document level (which may contain several topics). We employed representative metrics from both groups to provide a broader spectrum on the performance.

Specifically, we used six evaluation measures as the main metrics. They consist of four label-based measures: macro-F1, macro-Average Precision (AP), micro-F1, and micro-AP and two instance-based measures: instance-based F1 and accuracy (also stands for exact match ratio and the complement of zero one loss in this case). We further reported six evaluation measures that have been used to calculate the main metrics as additional metrics. They consist of four label-based measures: macro-Precision, macro-Recall, micro-Precision, and micro-Recall and two label-based measures: instance-based Precision and instance-based Recall. Their calculation formulas are summarized below.

### 3.5.1 Label-based measures

Label-based measures evaluate the multi-label classifier's performance separately on each label by calculating their true positive (TP), false positive (FP) and false negative (FN) on the test set. For the $j$-th label $y_j$, we calculated the following four metrics:

$$Precision_j = \frac{TP_j}{TP_j + FP_j}$$

$$Recall_j = \frac{TP_j}{TP_j + FN_j}$$

$$F1_j = \frac{2 \cdot Precision_j \cdot Recall_j}{Precision_j + Recall_j}$$

$$AP_j = \sum_n \frac{Recall_{j_n} - Recall_{j_{n-1}}}{Precision_{jn}}$$

F1 and AP are aggregated measures using both Precision and Recall in the calculation. AP is also a threshold-based measure which summarizes a Precision-Recall curve at each threshold (denoted as n in the formula).

To measure the overall metrics for all the labels, we calculated both macro-averaged (using unweighted averaging across labels) and micro-averaged scores (counting TP, FP and FN globally rather than at label level) for the labels.

### 3.5.2 Example-based measures

The example-based metrics evaluate the multi-label classifier's performance separately by comparing the predicted labels with the gold-standard labels for each test example. We focus on the following four metrics:

$$Precision = \frac{1}{p} \sum_{i=1}^{p} \frac{|Y_i \cap \hat{Y}_i|}{|\hat{Y}_i|} = \frac{1}{p} \sum_{i=1}^{p} \frac{TP}{TP + FP}$$

$$Recall = \frac{1}{p} \sum_{i=1}^{p} \frac{|Y_i \cap \hat{Y}_i|}{|Y_i|} = \frac{1}{p} \sum_{i=1}^{p} \frac{TP}{TP + FN}$$

$$F1 = \frac{2 \cdot Precision \cdot Recall}{Precision + Recall}$$

$$Accuracy = \frac{\text{\# correctly predicted instances}}{\text{\# all the instances}}$$

where p is the number of documents in the test set; $Y_i$ refers to the true label set for the i-th document in the test set; and $\hat{Y}_i$ refers to the predicted label set for the i-th document in the test set.

### 3.5.3 Statistic test and reporting standard

We repeated each model 10 times, reported the mean and max values of the repeats for each evaluation measure above, and conducted the Wilcoxon rank-sum test (Confidence Interval at 95%; one-tail) following previous studies [40, 47]

## 4 RESULTS AND DISCUSSIONS

### 4.1 Overall performance

TABLE 3 demonstrates the overall performance of the models on both datasets. As mentioned, we used six main metrics and reported their mean and max results (i.e., 12 evaluation measurement results). Out of these 12 measurement results, LITMC-BERT consistently achieved the highest results in 10 of them in the LitCovid BioCreative dataset and all the 12 in the HoC dataset. On average, its macro F1-score is about 10% higher than ML-Net in both datasets; the same applies to other measures such as macro-AP and accuracy. Compared with Binary BERT, its label-based measures are about 2% and 4% higher on the LitCovid BioCreative and HoC datasets, respectively. Its instance-based measures on the HoC dataset show a larger difference; e.g., its accuracy is up to 10% higher. The observations are similar when comparing LITMC-BERT with Linear BERT: e.g., its macro-F1 and accuracy are up to 2% and 4% higher on the HoC dataset, respectively. In terms of comparing Binary BERT with Linear BERT, Binary BERT achieved overall better performance on the LitCovid BioCreative dataset, which is consistent with the literature [6, 15], whereas Linear BERT achieved over better performance on the HoC dataset.

In addition, Fig. 4 shows the distribution of macro F1-scores of the models and the *P-values* of the Wilcoxon rank-sum test. On both datasets, LITMC-BERT consistently had a better macro-F1 score than ML-Net (*P-values* close to 0) and both Binary BERT and Linear BERT (*P-values* smaller or close to 0.001).

Further, comparing with the current SOTA results on the HoC dataset, LITMC-BERT also achieved higher performance. For LitCovid BioCreative, LITMC-BERT achieved better performance than the results reported by the challenge overview from 80 system submissions worldwide [37]. Existing studies on the HoC dataset used different measures and only reported one evaluation result (without repetitions to report the average performance or perform statistic tests). One study used instance-based F1 and reported that BlueBERT (base) and BlueBERT (large) achieved the highest instance-based F1 of 0.8530 and 0.8730, respectively, compared with other BERT variants [16]. In contrast, LITMC-BERT achieved a mean instance-based F1 of 0.9030 and a maximum instance-based F1 of 0.9169, consistently higher than the reported performance. Similarly, another study used micro-F1 on a slightly different version of the HoC dataset (this is different from other studies [16, 24]) and reported that PubMedBERT achieved the highest micro-F1 of 0.8232 [17]. The mean and maximum of micro-F1 of LITMC-BERT are 0.8648 and 0.8787, respectively. We manually examined the results and find that one possible reason is that the existing studies use the BERT model at sentence-level and then aggregate the predictions to the abstract-level for the HoC dataset [16]; this may ignore the inter-relations among sentences and cannot capture the context at abstract-level. In contrast, we directly applied the models at the abstract-level which overcomes the limitations.

## 4.2 Additional measures, label-specific results, and an ablation analysis

TABLE 4 provides additional measures to complement the main metrics. As mentioned, we reported the mean and maximum of six additional metrics (i.e., 12 in total). Out of these 12 additional measurement results, LITMC-BERT achieved the highest results in 7 of them in the LitCovid BioCreative dataset and 11 of them in the HoC dataset, which is consistent with the main measurement results in TABLE 3.

In addition, we further analyzed the performance of each individual label. Fig. 5 and Fig. 6 show F1s of each label in the LitCovid BioCreative and HoC datasets, respectively. Out of the seven labels in the LitCovid BioCreative dataset, LITMC-BERT had the highest F1 in four of them. Similarly, it had the highest F1 in seven out of 10 labels in the HoC dataset. The results also demonstrate that LITMC-BERT had much better performance for labels with low frequencies. For the LitCovid BioCreative dataset, its F1s are up to 9% and 6% higher for the Epidemic Forecasting (accounting for 1.64% of the testing set) and Transmission (5.12%) labels than Binary BERT and Linear BERT, respectively. For the HoC dataset, its F1s are also up to 6% and 5% higher for the Avoiding immune destruction (5.40%) and Enabling replicative immortality (5.71%) labels than Binary BERT and Linear BERT, respectively. This suggests that LITMC-BERT might be more robust to the class imbalance issue. Indeed, existing studies have demonstrated the class imbalance issue remains an open challenge for multi-label classifications and it is more difficult to improve the classification performance for rare classes [21], [22]. This is more evident that BERT-related models can already achieve F1-scores of close to or above 90% for labels with high frequencies on both datasets. Therefore, the performance of topics with low frequencies is arguably more critical.

Further, we performed an ablation analysis to quantify the effectiveness of the LITMC-BERT modules. Specifically, we compared the performance of LITMC-BERT without the Label Module, the Label Pair Module, or both (i.e., Linear BERT) using the same evaluation procedure above. TABLE 5 shows the results. Recall that Linear BERT uses the same BERT backbone and does not capture label-specific features or correlations between labels; therefore, we can directly compare the effectiveness of the Label Module and the Label Pair Module with Linear BERT. On both datasets, LITMC-BERT with both modules had the highest performance in all the measures. For instance, the Label Module increased the average macro-F1 by 2.1% and 0.5% in LitCovid BioCreative and HoC, respectively; the Label Pair Module increased the average macro-F1 by 0.7% and 0.5% in LitCovid BioCreative and HoC, respectively. Consistent observations are also shown in other metrics; for example, the Label Module increased the average macro-AP by 2.5% and 1.5% in LitCovid BioCreative and HoC, respectively. This suggests that the two modules complement to each other and combining both is effective. In addition, removing either module dropped the performance in both datasets; removing both of them had the lowest performance on average. This suggests both modules are effective. Also, the results suggest that the Label Module is more effective in the LitCovid BioCreative dataset (e.g., the maco-F1 is reduced by up to 2% if removing the Label Module) whereas the Label Pair Module is more effective in the HoC dataset (e.g., the maco-F1 is reduced by up to 1% if removing the Label Pair Module).

## 4.3 Generalization and efficiency analysis

The above evaluations show that LITMC-BERT achieved consistently better performance on both datasets. We also further evaluated its generalization and efficiency in the LitCovid production environment. While transformer-based models have achieved SOTA results in many applications, their inference time is significantly longer than other types of models [48]. It is thus important to measure its efficiency in practice. Reducing inference time is also critical to the LitCovid curation pipeline, which may have thousands of articles to curate within a day (the peak was over 2,500 articles in a single day) [6]. A random sample of 3,000 articles in LitCovid was collected between October and December, 2021, which was independent to the training set. We used it as an external validation set and measured the accuracy and efficiency of these models. As mentioned, Binary BERT was used in LitCovid for topic annotations. We used a single processor on CPU with a batch size of 128, which is consistent with the LitCovid production setting, and tracked the inference time accordingly. TABLE 6 details the performance. LITMC-BERT achieved the best performance in all the accuracy-related measures, took ~18% of the prediction time of Binary BERT, and was only 0.05 sec/doc slower than Linear BERT as the trade-off. Note that Binary BERT was previously used in the LitCovid production. It took about 3.4 seconds on average to predict topics for an article. Note that this does not include overhead time (e.g., switching into other automatic curation tasks) and post-processing time (e.g., sorting the probabilities and showing the related articles for manual review). Therefore, just predicting the topics for a large batch of articles may take over an hour, delaying the daily curation of LitCovid. In contrast, it only takes LITMC-BERT about 0.5 seconds on average for inference, which accounts for ~15% of the time used by Binary BERT. We have employed LITMC-BERT into the LitCovid production system given its superior performance on both effectiveness and efficiency.

## 4.4 Limitations and future work

While LITMC-BERT achieved the best overall performance in both datasets compared with other competitive baselines, it does have certain limitations that we plan to address in the future. First, from the method level, it still relies on transfer learning from BERT backbones given the scale of multi-label classification datasets in biomedical literature. In contrast, some methods used other domains include label clustering [49] and label graph attentions [50]. We plan to investigate these methods and quantify whether they are effective for biomedical literature. Second, it has more hyperparameters to tune (such as finding the optimal label pair threshold) compared with other straightforward BERT-based models. It would be better to incorporate dynamic modules that learn these hyperparameters adaptively. Third, the Label Pair Module focuses only on co-occurred labels which may miss more complex scenarios such as labels in n-ary relations.

# 5 CONCLUSION

In this paper, we propose a novel transformer-based multi-label classification method on biomedical literature, LITMC-BERT. Compared to the existing multi-label classification methods in biomedical literature, it captures label-specific features and also captures the correlations between label pairs. The multi-task training approach also makes it more efficient than binary models. LITMC-BERT achieved the highest overall performance on two datasets than three baselines. Also, it only takes ~18% of the inference time taken by the previous best model for COVID-19 literature. LITMC-BERT has been employed in the LitCovid production system for more sustainability and effectiveness. We plan to further improve the method such that it is more dynamic and capable of handling more complex relations among labels and further quantify its effectiveness on multi-label classification tasks beyond biomedical literature (such as clinical notes).

## ACKNOWLEDGMENT

The authors thank Dr. Robert Leaman for proofreading the manuscript. This research is supported by the NIH Intramural Research Program, National Library of Medicine.

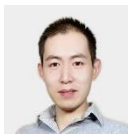

Qingyu Chen received Ph.D. in Biomedical Informatics from the University of Melbourne. He is currently a research fellow at the BioNLP lab, National Center for Biotechnology Information (NCBI) at the National Library of Medicine (NLM), and the lead instructor of text mining courses at the Foundation for Advanced Education in the Sciences. His research interests include biomedical text mining, medical image analytics, and biocuration. Dr. Chen has published over 30 first-authored papers and 50 papers in total. He serves as the Associate Editor of Frontiers in Digital Health, ECR editor of Applied Clinical Informatics and PC member for the IEEE International Conference on Healthcare Informatics and ACL BioNLP workshop.

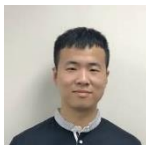

Jingcheng Du received Ph.D. in health informatics from The University of Texas Health Science Center at Houston (UTHealth), TX, USA. He is an assistant professor in health informatics at UTHealth School of Biomedical Informatics. His research interest include machine learning, biomedical natural language processing and knowledge representation.

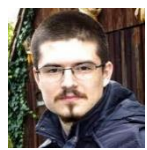

Alexis Allot received his PhD in Bioinformatics in 2015 from the University of Strasbourg, France. He then worked at EMBL/EBI as Bioinformatician and is now working as postdoctoral fellow at NIH. His research interests include biomedical text mining, data mining and web development.

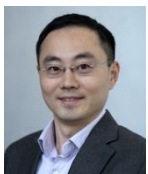

Zhiyong Lu received the Ph.D. degree in bioinformatics from the School of Medicine, University of Colorado in 2007. Dr. Lu is currently Deputy Director of Literature Search, National Center for Biotechnology Information (NCBI) at the National Library of Medicine (NLM), directing R&D efforts of improving literature searches such as PubMed and LitCovid. He is also an NIH Senior Investigator (with early tenure), leading the Natural Language Processing (NLP) and Machine Learning research at NLM/NIH. Dr. Lu has published over 200 scientific articles and books. He serves as Associate Editor for the journals Bioinformatics and Artificial Intelligence in Medicine. Dr. Lu is an elected fellow of the American College of Medical Informatics.

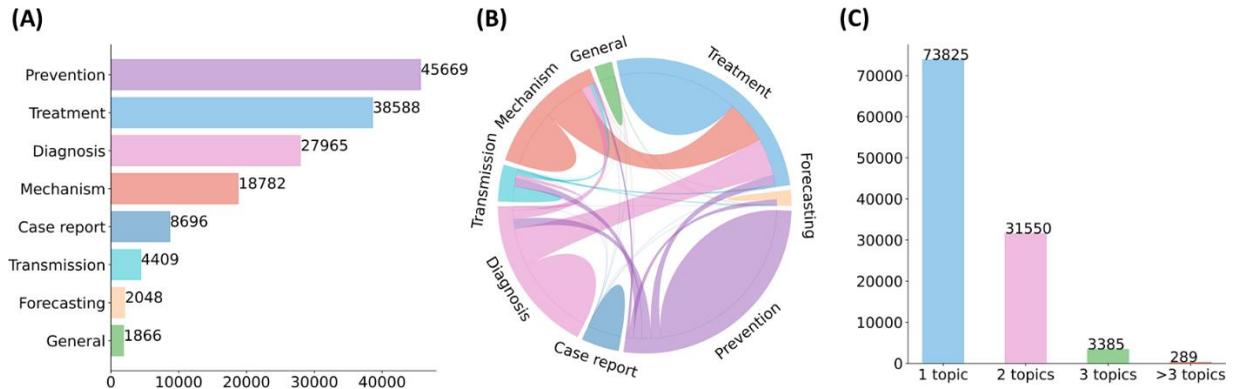

Fig. 1. Characteristics of topic annotations in LitCovid up to September 2021. (A) shows the frequencies of topics; (B) demonstrates topic co-occurrences; and (C) illustrates the distributions of the number of topics assigned per document. The figure is adapted from [37].

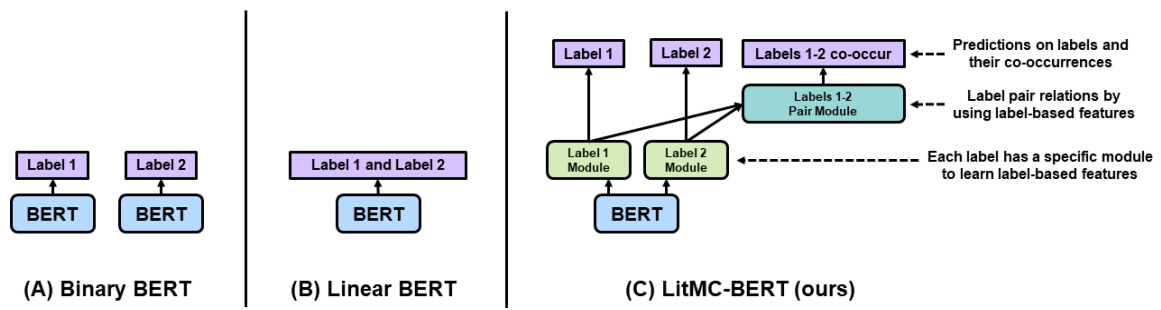

Fig. 2.. An overview of BERT-based multi-classification models for biomedical literature using an example of classifying two labels (Labels 1 and 2). (A): Binary BERT: train a BERT model for each label; (B) Linear BERT: train a shared BERT model and output all the label probabilities at once; (C): LitMC-BERT (our proposed approach): train a shared BERT model, capture label-based features (Label Module) and models pair relations (Label Pair Module), and also predict both labels and their co-occurrences via co-training.

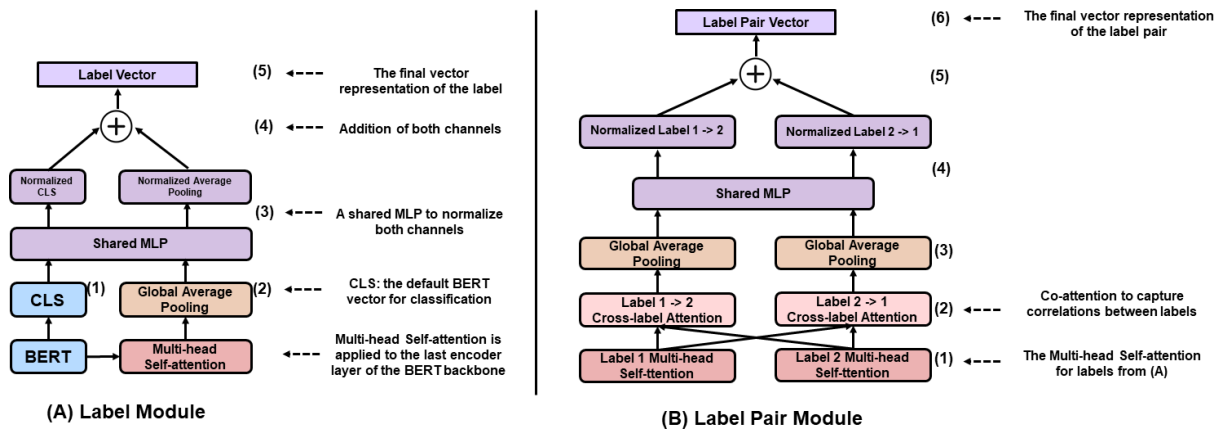

Fig. 3. The illustration of the Label Module and Label Pair Module. MLP: multi-layer perceptron. The detailed hyperparameters are provided in TABLE 2.

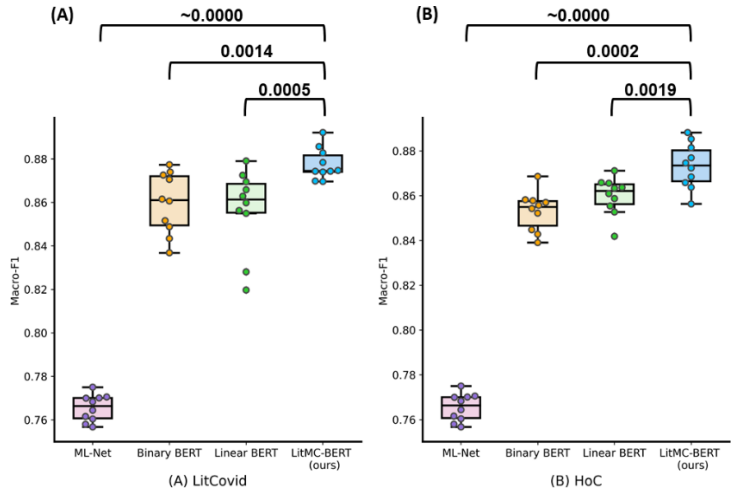

Fig. 4. The distributions of macro-F1s for each model on the LitCovid BioCreative (A) and HoC datasets (B). Each model was repeated 10 times and the Wilcoxon rank sum test (Confidence Interval at 95%; one-tail) was performed. The *P-values* are shown in the figure.

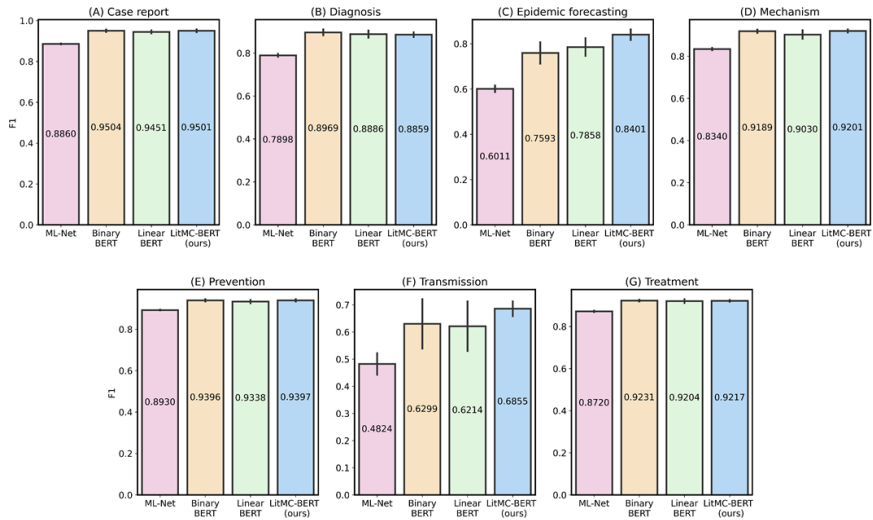

Fig. 5. The performance (F1) of the methods for each label in the LitCovid dataset.

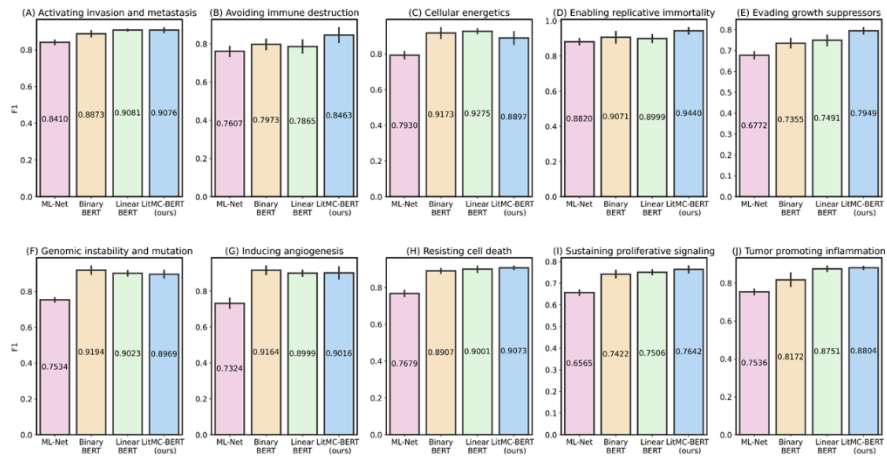

Fig. 6. The performance (F1) of the methods for each label in the HoC dataset.

TABLE 1

CHARACTERISTICS OF THE EXPERIMENT DATASETS. #ARTICLES: THE NUMBER OF ARTICLES; LABEL (%): THE PROPORTION OF THE ARTICLES WITH A SPECIFIC LABEL; SOME LABEL NAMES OF THE HOC DATASET ARE SHORTENED FOR REPRESENTATION PURPOSE.

| | Train | | Valid | | Test | | All | |
|---|---|---|---|---|---|---|---|---|
| | #Articles | Label (%) | #Articles | Label (%) | #Articles | Label (%) | #Articles | Label (%) |
| **LitCovid BioCreative (7 labels)** | 24,960 | - | 6,239 | - | 2,500 | - | 33,699 | - |
| Case Report | 2,063 | (8.27%) | 482 | (7.73%) | 197 | (7.88%) | 2,742 | (8.14%) |
| Diagnosis | 6,193 | (24.81%) | 1,546 | (24.78%) | 722 | (28.88%) | 8,461 | (25.11%) |
| Epidemic Forecasting | 645 | (2.58%) | 192 | (3.08%) | 41 | (1.64%) | 878 | (2.61%) |
| Mechanism | 4,438 | (17.78%) | 1,073 | (17.2%) | 567 | (22.68%) | 6,078 | (18.04%) |
| Prevention | 11,102 | (44.48%) | 2,750 | (44.08%) | 926 | (37.04%) | 14,778 | (43.85%) |
| Transmission | 1,088 | (4.36%) | 256 | (4.1%) | 128 | (5.12%) | 1,472 | (4.37%) |
| Treatment | 8,717 | (34.92%) | 2,207 | (35.37%) | 1,035 | (41.4%) | 11,959 | (35.49%) |
| **Hoc (10 labels)** | 1,108 | - | 157 | - | 315 | - | 1,580 | - |
| Activating invasion & metastasis | 199 | (17.96%) | 35 | (22.29%) | 57 | (18.1%) | 291 | (18.42%) |
| Avoiding immune destruction | 77 | (6.95%) | 14 | (8.92%) | 17 | (5.40%) | 108 | (6.84%) |
| Cellular energetics | 76 | (6.86%) | 10 | (6.37%) | 19 | (6.03%) | 105 | (6.65%) |
| Enabling replicative immortality | 82 | (7.40%) | 15 | (9.55%) | 18 | (5.71%) | 115 | (7.28%) |
| Evading growth suppressors | 174 | (15.7%) | 22 | (14.01%) | 46 | (14.60%) | 242 | (15.32%) |
| Genomic instability & mutation | 239 | (21.57%) | 24 | (15.29%) | 70 | (22.22%) | 333 | (21.08%) |
| Inducing angiogenesis | 97 | (8.75%) | 15 | (9.55%) | 31 | (9.84%) | 143 | (9.05%) |
| Resisting cell death | 302 | (27.26%) | 45 | (28.66%) | 83 | (26.35%) | 430 | (27.22%) |
| Sustaining proliferative signal | 338 | (30.51%) | 41 | (26.11%) | 83 | (26.35%) | 462 | (29.24%) |
| Tumor promoting inflammation | 162 | (14.62%) | 24 | (15.29%) | 54 | (17.14%) | 240 | (15.19%) |

TABLE 2

HYPERPARAMETERS OF THE METHODS. SEQ LEN: SEQUENCE LENGTH. BIOBERT: WE USED BIOBERT V1.0; MLP: MULTILAYER PERCEPTRON WITH THREE HIDDEN LAYERS, EACH WITH 512, 256, AND 128 HIDDEN UNITS, RESPECTIVELY.

| | ML-Net | | Binary BERT | | Linear BERT | | LITMC-BERT (ours) | |
|---|---|---|---|---|---|---|---|---|
| **Shared hyperparameters** | | | | | | | | |
| Max seq len | 2000 characters | | 512 tokens | | 512 tokens | | 512 tokens | |
| Backbone | ELMO | | BioBERT | | BioBERT | | BioBERT | |
| Batch size | 16 | | 16 | | 16 | | 16 | |
| Learning rate | 1e-3 | | 5e-2 | | 5e-2 | | 5e-2 | |
| Activation function | ReLU | | Sigmoid | | Sigmoid | | Sigmoid | |
| Loss function | Log-sum-exp pairwise | | Cross-entropy | | Cross-entropy | | Label predictions | Cross-entropy |
| | | | | | | | Pair predictions | Focal loss |
| **Specific hyperparameters** | RNN units | 50 | Early stop | 2 | Early stop | 2 | Early stop | 2 |
| | Attention units | 50 | | | | | MLP units (3 layers) | 512, 256, 128 |
| | | | | | | | Multi-head number | 16 |
| | | | | | | | Label pair threshold | 0.40 |
| | | | | | | | Auxiliary task weight | 0.25 |

TABLE 3

THE OVERALL PERFORMANCE (MAIN EVALUATION MEASURES) OF THE METHODS ON THE LITCOVID AND HOC DATSETS. *: THE REPORTED RESULTS WAS ON A SLIGHTLY DIFFERENT VERSION OF THE HOC DATASET.

| | Label-based measures | | | | | | | | Instance-based measures | | | |
|---|---|---|---|---|---|---|---|---|---|---|---|---|
| | Macro-F1 | | Macro-AP | | Micro-F1 | | Micro-AP | | F1 | | Accuracy | |
| | Mean | Max | Mean | Max | Mean | Max | Mean | Max | Mean | Max | Mean | Max |
| **LitCovid BioCreative** | | | | | | | | | | | | |
| ML-Net | 0.7655 | 0.7750 | - | - | 0.8437 | 0.8470 | - | - | 0.8678 | 0.8706 | 0.7019 | 0.7108 |
| Binary BERT | 0.8597 | 0.8773 | 0.7825 | 0.8059 | **0.9132** | 0.9186 | **0.8557** | 0.8655 | 0.9278 | 0.9330 | 0.7984 | 0.8120 |
| Linear BERT | 0.8569 | 0.8791 | 0.7796 | 0.8066 | 0.9067 | 0.9163 | 0.8461 | 0.8607 | 0.9254 | 0.9341 | 0.7915 | 0.8072 |
| LitMC-BERT (ours) | **0.8776** | **0.8921** | **0.8048** | **0.8223** | 0.9129 | **0.9212** | 0.8553 | **0.8663** | **0.9314** | **0.9384** | **0.8022** | **0.8188** |
| **Hoc** | | | | | | | | | | | | |
| ML-Net | 0.7618 | 0.7665 | - | - | 0.7449 | 0.7560 | - | - | 0.7931 | 0.8003 | 0.4990 | 0.5429 |
| Binary BERT | 0.8530 | 0.8686 | 0.7581 | 0.7811 | 0.8453 | 0.8583 | 0.7368 | 0.7568 | 0.8733 | 0.8850 | 0.6251 | 0.6476 |
| Linear BERT | 0.8599 | 0.8711 | 0.7690 | 0.7875 | 0.8554 | 0.8637 | 0.7547 | 0.7670 | 0.8941 | 0.9018 | 0.6695 | 0.6857 |
| LitMC-BERT (ours) | **0.8733** | **0.8882** | **0.7894** | **0.8118** | **0.8648** | **0.8787** | **0.7697** | **0.7905** | **0.9036** | **0.9169** | **0.6854** | **0.7270** |
| **Reported SOTA performance on Hoc** | | | | | | | | | | | | |
| BlueBERT (base) | - | - | - | - | - | - | - | - | - | 0.8530 | - | - |
| BlueBERT (large) | - | - | - | - | - | - | - | - | - | 0.8730 | - | - |
| PubMedBERT | - | - | - | - | - | 0.8232* | - | - | - | - | - | - |

TABLE 4

ADDITIONAL EVALUATION MEASURES OF THE METHODS ON THE LITCOVID AND HOC DATSETS.

| | Label-based measures | | | | | | | | Instance-based measures | | | |
|---|---|---|---|---|---|---|---|---|---|---|---|---|
| | Macro-Precision | | Macro-Recall | | Micro-Precision | | Micro-Recall | | Precision | | Recall | |
| | Mean | Max | Mean | Max | Mean | Max | Mean | Max | Mean | Max | Mean | Max |
| **LitCovid BioCreative** | | | | | | | | | | | | |
| ML-Net | 0.8364 | 0.8559 | 0.7309 | 0.7632 | 0.8756 | 0.8827 | 0.8142 | 0.8227 | 0.8849 | 0.8901 | 0.8514 | 0.8591 |
| Binary BERT | 0.9103 | **0.9498** | 0.8350 | 0.8690 | 0.9304 | **0.9448** | 0.8969 | **0.9173** | 0.9349 | 0.9408 | 0.9210 | **0.9381** |
| Linear BERT | 0.9071 | 0.9388 | 0.8354 | 0.8701 | 0.9276 | 0.9396 | 0.8870 | 0.9093 | 0.9368 | 0.9443 | 0.9143 | 0.9308 |
| LitMC-BERT (ours) | **0.9131** | 0.9226 | **0.8574** | **0.8814** | **0.9313** | 0.9366 | 0.8952 | 0.9145 | **0.9418** | **0.9480** | **0.9212** | 0.9355 |
| **Hoc** | | | | | | | | | | | | |
| ML-Net | 0.7949 | 0.8356 | 0.7389 | 0.7622 | 0.7667 | 0.8053 | 0.7253 | 0.7448 | 0.8045 | 0.8296 | 0.7826 | 0.8013 |
| Binary BERT | 0.8471 | 0.8644 | **0.8661** | 0.8834 | 0.8363 | 0.8562 | 0.8548 | 0.8724 | 0.8565 | 0.8735 | 0.8909 | 0.9095 |
| Linear BERT | 0.8772 | 0.8930 | 0.8475 | 0.8630 | 0.8614 | 0.8812 | 0.8496 | 0.8619 | 0.8929 | 0.9116 | 0.8955 | 0.9024 |
| LitMC-BERT (ours) | **0.8868** | **0.8983** | 0.8641 | **0.8910** | **0.8718** | **0.8938** | **0.8582** | **0.8787** | **0.9035** | **0.9148** | **0.9038** | **0.9193** |

TABLE 5

ABLATION ANALYSIS ON THE EFFECTIVENESS OF THE LABEL MODULE AND LABEL PAIR MODULES. NOTE THAT 'NO BOTH MODULES' IS LINEAR BERT.

| | Label-based measures | | | | | | | | Instance-based measures | | | |
|---|---|---|---|---|---|---|---|---|---|---|---|---|
| | Macro-F1 | | Macro-AP | | Micro-F1 | | Micro-AP | | F1 | | Accuracy | |
| | Mean | Max | Mean | Max | Mean | Max | Mean | Max | Mean | Max | Mean | Max |
| **LitCovid BioCreative** | | | | | | | | | | | | |
| LitMC-BERT (ours) | **0.8776** | **0.8921** | **0.8048** | **0.8223** | **0.9129** | **0.9212** | **0.8553** | **0.8663** | **0.9314** | **0.9384** | **0.8022** | **0.8188** |
| -No Label Module | 0.8635 | 0.8741 | 0.7868 | 0.7984 | 0.9071 | 0.9117 | 0.8474 | 0.8528 | 0.9258 | 0.9288 | 0.7942 | 0.8068 |
| -No Label Pair | 0.8752 | 0.8838 | 0.8018 | 0.8121 | 0.9116 | 0.9166 | 0.8536 | 0.8606 | 0.9298 | 0.9326 | 0.8011 | 0.8104 |
| -No both modules | 0.8569 | 0.8791 | 0.7796 | 0.8066 | 0.9067 | 0.9163 | 0.8461 | 0.8607 | 0.9254 | 0.9341 | 0.7915 | 0.8072 |
| **Hoc** | | | | | | | | | | | | |
| LitMC-BERT (ours) | **0.8733** | **0.8882** | **0.7894** | **0.8118** | **0.8648** | **0.8787** | **0.7697** | **0.7905** | **0.9036** | **0.9169** | **0.6854** | **0.7270** |
| -No Label Module | 0.8691 | 0.8796 | 0.7860 | 0.8027 | 0.8580 | 0.8638 | 0.7602 | 0.7717 | 0.8984 | 0.9046 | 0.6803 | 0.7016 |
| -No Label Pair | 0.8644 | 0.8803 | 0.7762 | 0.8005 | 0.8590 | 0.8700 | 0.7618 | 0.7807 | 0.8998 | 0.9071 | 0.6756 | 0.7206 |
| -No both modules | 0.8599 | 0.8711 | 0.7690 | 0.7875 | 0.8554 | 0.8637 | 0.7547 | 0.7670 | 0.8941 | 0.9018 | 0.6695 | 0.6857 |

TABLE 6
PERFORMANCE EVALUATION ON AN INDEPENDENT SAMPLE OF 3,000 ARTICLES IN LITCOVID. TIME WAS MEASURED IN THE PRODUCTION ENVIRONMENT WHERE ONLY CPUS ARE AVAILABLE.

| | Label-based measures | | | | Instance-based measures | | Time measures in the production environment (in sec) |
|---|---|---|---|---|---|---|---|
| | **Macro-F1** | **Macro-AP** | **Micro-F1** | **Micro-AP** | **F1** | **Accuracy** | |
| Binary BERT | 0.8116 | 0.7251 | 0.8570 | 0.7733 | 0.8719 | 0.6327 | 8912.22 (2.97 sec/doc) |
| Linear BERT | 0.8299 | 0.7449 | 0.8539 | 0.7701 | 0.8759 | 0.6354 | **1536.24** **(0.51 sec/doc)** |
| LitMC-BERT (ours) | **0.8413** | **0.7690** | **0.8628** | **0.7813** | **0.8862** | **0.6512** | 1673.98 (0.56 sec/doc) |